**Title page**

## Title

AI-based characterization of lymphoid structures and the tumor microenvironment reveals lower mature tertiary lymphoid structures, lymph-node follicles, and plasma cells after neoadjuvant therapy in lung squamous cell carcinoma

## Authors

Arne Schmidt[1], Justin Nieder[2,3], Cornelius Boehm[1], Julika Ribbat-Idel[1,4], Marius Teodorescu[1], Marie-Lisa Eich[5], Simon Schallenberg[5], Gabriel Dernbach[1,5], Katja Lingelbach[1], Miriam Haegele[1], Timo Milbich[1], Adam Narai[1], Maria Stoelben[1], Hans Pinckaers[1], Lukas Ruff[1], Kai Standvoss[1], Christian Watermann[2], Christiane Kuempers[2,4], Sven Perner[7], Verena Sailer[2,4], Jasmin Tiebach[6], Torsten Goldmann[6], Sebastian Marwitz[6], Johannes Braegelmann[8], Frederick Klauschen[1,9,10,11,12], Marija Pezer[1], Rosemarie Krupar[1,4]

## Affiliations

[1] Aignostics GmbH, Berlin, Germany

[2] Institute of Pathology, University Hospital Schleswig-Holstein (UKSH), Campus Luebeck, Germany

[3] Schoen Clinic Group Neustadt, Neustadt, Germany

[4] University of Luebeck, Luebeck, Germany

[5] Institute of Pathology, Charité – Universitätsmedizin Berlin, Berlin, Germany

[6] Research Center Borstel, Leibniz Lung Center, Borstel, Germany

[7] PATHORA Institute of Pathology and Tissue Medicine, Reutlingen, Germany

[8] Medical Clinic III for Oncology, Hematology, Immune-Oncology and Rheumatology, University Hospital Bonn (UKB), Bonn, Germany

[9] BIFOLD – Berlin Institute for the Foundations of Learning and Data, Berlin, Germany

[10] Institute of Pathology, Ludwig-Maximilians-University, Munich, Germany

[11] German Cancer Consortium (DKTK), German Cancer Research Center (DKFZ), Munich Partner Site, Heidelberg, Germany

[12] Bavarian Cancer Research Center (BZKF), Munich Partner Site, Munich, Germany

## Running title

Neoadjuvant therapy-associated differences in TLS, SLS and TME of LUSC

## Additional information

Corresponding author: Rosemarie Krupar; Aignostics GmbH, Alt-Moabit 73/73A, 10555 Berlin; Email: rosemarie.krupar@aignostics.com

## Conflict of interest disclosure

## Additional information

Manuscript word count: 7409 including title

Number of figures: 4

Number of tables: 1

**Abstract**

Mature, germinal centre–bearing tertiary lymphoid structures (TLS) and mature lymph node follicles (secondary lymphoid structures/SLS) sustain the plasma-cell humoral response that shapes outcome and immunotherapy benefit in lung cancer. How this loco-regional axis differs between neoadjuvant-treated and therapy-naïve residual tumors is poorly defined. We analyzed H&E whole-slide images of the resected tumor and paired proximal and distal lymph nodes from matched neoadjuvant-treated (neoLUSC, n=20) and therapy-naïve (naivLUSC, n=20) lung squamous cell carcinoma (LUSC) patients. Aignostics Atlas H&E Tissue Analysis (HETA) models quantified single-cell densities in different tissue compartments, and cell–cell co-localizations. A dedicated deep-learning model, validated against expert pathologist annotations, detected TLS and SLS and graded each as mature or immature. Prognostic value of cell-level readouts was tested in an independent LUSC cohort (n=458, 412 therapy-naïve and 46 treated patients). In neoLUSC patients stromal and carcinoma immune cell densities were lower, most pronounced for the humoral axis: Stromal plasma cells were 72% (p=0.004) and stromal lymphocytes 48% (p=0.015) lower in neoLUSC tumors compared to naivLUSC tumors. This difference was not observed for macrophages and granulocytes. With regard to TLS the difference was selective for maturity: Total and immature TLS did not differ between cohorts, whereas mature TLS in the tumor (p=0.034) and the tumor-distal lymph node (p=0.003) were lower in neoLUSC patients (naïve vs chemotherapy p=0.018, naïve vs chemoradiation p=0.007). Spatially, the peri-TLS plasma cell to tumor lymphocyte association present in naivLUSC tumors (ρ=0.54, p=0.017) was not observed in neoLUSC tumors (ρ=0.07). In the validation cohort, the cell-level readouts that were lower in treated tumors were prognostic: Stromal plasma cell density (HR 0.69, p=0.011) and lymphocyte–plasma cell co-localization (HR 0.66, p=0.004) predicted longer survival, while neoadjuvant status was adverse (HR 1.50, p=0.026). Neoadjuvant therapy in LUSC thus seems to be associated with fewer mature lymphoid structures and with weaker coupling of the humoral compartment to the tumor rather than with uniformly lower immune readouts, most pronounced after chemoradiation. The readouts that were lower in treated tumors are prognostic in an independent cohort, nominating the stromal plasma cell/co-localization axis as a candidate biomarker.

## Introduction

Lung cancer remains the most commonly diagnosed malignancy and the leading cause of cancer-related death worldwide, accounting for approximately 2.5 million new cases and 1.8 million deaths in 2022 [1]. Non–small cell lung cancer (NSCLC) comprises roughly 80–85% of cases, with lung squamous cell carcinoma (LUSC) representing the second most common subtype after adenocarcinoma [2]. Because most patients present with advanced disease, long-term outcomes are poor, with an overall five-year relative survival of approximately 29% [3]. Immune-checkpoint inhibitors (ICI) directed against the PD-1/PD-L1 axis have become a cornerstone of NSCLC treatment, yet 50–80% of patients derive no durable benefit, and PD-L1 expression alone remains an imperfect predictor of response [4,5]. This has intensified interest in the broader architecture of the tumor immune microenvironment (TIME) as a determinant of antitumor immunity.

Beyond individual immune-cell densities, the spatial organization of the TIME carries independent prognostic and predictive value [6]. Tertiary lymphoid structures (TLS) — ectopic lymphoid aggregates that recapitulate the architecture and function of secondary lymphoid organs — are associated with favorable prognosis and improved ICI response in NSCLC [7]. Critically, it is the maturation state of these structures, rather than their mere presence, that determines their clinical relevance: a germinal center (GC)–bearing, mature TLS sustains the B-cell reactions that drive plasma-cell differentiation and the production of tumor-specific antibodies [8,9]. This plasma-cell output fuels an intratumoral immunity cycle [10], and mature TLS predict ICI efficacy independently of PD-L1 expression and CD8+ T-cell density [11].

This antigen-driven response is not confined to the tumor. Antigen presentation and B-cell maturation classically occur within the follicles and germinal centers of draining lymph nodes, the systemic arm of the humoral response [12], and intratumoral TLS maturation is coupled to the maturation state of these regional nodes in lung cancer [13,14]. Together, the intratumoral (TLS) and regional (lymph node follicle/GC) compartments form a spatially linked axis whose disruption could blunt the entire plasma-cell–driven response.

Neoadjuvant chemotherapy or chemoradiation, increasingly combined with a PD-L1 inhibitor, is now standard of care for resectable stage III NSCLC [15]. On the one hand, cytotoxic therapy can actively promote antitumor immunity: Chemotherapy- and radiation-induced immunogenic cell death releases tumor antigens and damage-associated molecular patterns that activate dendritic cells and prime $CD8^+$ T-cell responses [16]. Radiation in particular can engage cytosolic DNA sensing via the cGAS–STING axis and type-I interferon signalling, reshaping the tumor microenvironment (TME) toward a more immunogenic, T-cell–inflamed state [17,18]. Yet these modalities exert pronounced suppressive effects on the very B-cell and germinal-center compartments that underlie mature lymphoid structures, including lymphocyte depletion [19], loss of CD20+ B cells and TLS density in recurrent squamous tumors after chemoradiation [20], and impaired TLS maturation with germinal-center attrition [8].

How these opposing effects ultimately resolve in LUSC is difficult to read from existing data, because the local and regional arms of the response have mostly been examined one compartment at a time, and rarely with the scalable, whole-slide quantification needed to relate the intratumoral and lymph node compartments within the same patients. We therefore set out to resolve, across the full tumor to lymph node axis, if neoadjuvant chemotherapy and radiotherapy are associated with differences in the local TME, the intratumoral TLS, and the regional lymph node follicles. We focused on whether treatment is associated with differences in the maturation of lymphoid structures and in the coupling between the tumor and lymph node compartments, and how these differences relate to prognosis. To do so, we applied AI-based whole-slide segmentation and cell classification, built on a pathology foundation model, to generate quantitative, cell-resolved and spatially explicit readouts, an approach that scales the analysis well beyond manual assessment and renders the local–regional circuit measurable as a whole. We examined these readouts in a matched cohort of neoadjuvant treated (neoLUSC) and treatment-naïve surgical (naivLUSC) LUSC patients, each case paired per patient with a tumor-proximal and a tumor-distal lymph node, so that loco-regional differences between treated and therapy-naïve patients in the mature-lymphoid and plasma cell compartments that drive the antitumor humoral response could be compared along a single tumor-to-lymph node axis.

## Materials and methods

### Study cohorts

Two independent LUSC cohorts were analyzed in this study:

1. Discovery cohort: The first, whole slide image (WSI)-based, discovery cohort was created from the archives of the Institute of Pathology, UKSH Campus Lübeck, and the Pathology of the Research Center Borstel (Leibniz Lung Center) consisting of 40 patients, who were diagnosed with LUSC between 2009 - 2020 and treated by oncological tumor resection with locoregional lymphadenectomy. Overall survival data, patient characteristics and social histories were provided by the University Hospital Schleswig-Holstein, Campus Lübeck, the LungenClinic Grosshansdorf and the Institute for Cancer Epidemiology of the University Lübeck. The neoadjuvantly treated neoLUSC subgroup (n=20) comprised patients who had received neoadjuvant chemotherapy and/or radiotherapy prior to resection, including patients who died during neoadjuvant therapy and for whom autopsy tissue was available. The therapy-naïve naivLUSC subgroup (n=20) was restricted to patients who had received a surgical resection alone with no prior neoadjuvant conservative treatment and served as a matched control. Matching was performed manually using pathologically assessed TNM status, UICC stage, sex, and age at diagnosis. Cases were stratified by TNM status and assigned proportionally to the neoLUSC distribution, with ties resolved by the next-ranked criterion. The use of patient data and tumor material was approved by the Ethics Committee of the University of Lübeck (vote of 08 November 2016, AZ 16-277), and all material was fully anonymized. NeoLUSC patients were restricted to Junker regression grade IIa (>10% vital tumor), ensuring both a demonstrable therapy response and sufficient evaluable tumor tissue. Each resection was paired with a tumor-proximal (N1) and a tumor-distal (N2/3) tumor-free lymph node of the same patient, allowing the local TME and its draining nodal compartments to be examined along a single tumor-to-node axis within each patient. Paired tumor-free lymph nodes were evaluable for 37 of 40 patients proximally (N1) and 39 of 40 distally (N2/3). Four lymph node slides were excluded due to out of focus scans. The excluded lymph nodes (three proximal and one distal) were all from the naivLUSC cohort. Neoadjuvant modality comprised chemotherapy (n=11), chemoradiotherapy (n=5), and radiotherapy alone (n=2), with the modality undocumented in two patients. Mean age was 61.2 years (SD 7.1) in the neoLUSC group and 69.1 years (SD 6.9) in the naivLUSC group. The age difference was significant ($p = 0.001$). TNM staging followed the 8th edition; pT, pN, R-status, and UICC-stage (IA3–IVA). The distributions of all clinical and pathological parameters per group are given in **Table 1**.

2. The validation cohort: The second cohort was a tumor microarray (TMA)-based TME-focused validation cohort. It was created as a subset from a large, independent, well-characterized and previously described non-small cell lung cancer cohort [21] and included 458 LUSC patients, of which 46 patients had received neoadjuvant chemotherapy and/or radiation prior to resection and 412 patients had received resection without prior therapy.

### Sample selection and WSI processing

For the discovery cohort three representative H&E-stained sections were selected per patient by a board-certified pathologist (R. K.): (1) Tumor tissue with peritumoral parenchyma, (2) a tumor-free lymph node proximal to the primary within the N1 region (stations 10, 11, 12), and (3) a tumor-free lymph node distal to the primary within the N2/3 region (stations 4, 5, 6). Conventional H&E stains were prepared following established protocols and digitized on a Ventana BenchMark Ultra (Roche, Basel, Switzerland). WSIs were visualized in Aignostics' in house slide viewer

Portal. Annotations for model training, model performance measurements and downstream analyses were also executed in Portal by four board-certified anatomical pathologists (M.-L. E., J. R.-I., M. T., R. K.).

**AI-based analysis of the tumor microenvironment**

The TME analysis used Aignostics Atlas H&E Tissue Analysis (HETA) model outputs [22], executing a sequential per-slide workflow : Tissue QC, tissue segmentation, cell detection and cell classification followed by quantification of spatial readouts. Tissue QC identified valid tissue regions. Valid tissue was segmented into seven tissue types (carcinoma, stroma, epithelium, vessel, blood, necrosis, and other). Cells were detected and classified within valid tissue, excluding necrosis and blood, into nine cell types (carcinoma cell, epithelial cell, endothelial cell, fibroblast, granulocyte, lymphocyte, macrophage, plasma cell, and other), with nuclear morphometric features recorded per cell and each cell assigned to its tissue region. Whole tumor regions (WTR), defined as the outermost boundary enclosing all carcinoma areas, as well as interspaced non-carcinoma TME areas within 500 µm distance of a carcinoma area, were manually annotated by a board-certified anatomical pathologist for each tumor slide. Peri- or intratumoral lymph nodes were excluded from WTR annotations. Additionally, board-certified anatomical pathologists annotated lymphoid areas for each lymph node slide for density normalization and cell level readouts. All model predictions were quantified within the WTR regions. Readouts comprised per-region cell counts, percentages, and densities by cell class, per–tissue-type area and morphometric summaries.

**AI-based analysis of tertiary lymphoid structures and lymphoid follicles**

A TLS segmentation model built on Aignostics' Atlas foundation model backbone [23] was trained to predict three classes (no TLS, immature TLS, mature TLS) using annotations from 29 slides, then evaluated on a held-out set of 10 annotated tumor slides. On raw pixel-level predictions, performance was high for the 'no TLS' class (F1 0.99) and moderate for the 'immature TLS' and 'mature TLS' classes (F1 0.73 and 0.53, respectively). External model validation on an independent urothelial carcinoma hold-out set (10 slides) confirmed reliable 'no TLS' detection (F1 0.99) with more modest performance for the 'immature TLS' and 'mature TLS' classes (F1 0.62 and 0.50, respectively) for raw-pixel predictions.

Predictions were then post-processed by excluding those with fewer than 50 lymphocytes, a threshold retained from in house work and consistent with literature convention, and merging spatially close predictions [19]. A maturity threshold, tuned on the training set, classified a TLS as mature when >25% of its area was predicted mature, the best-performing of the tested thresholds (>0%, >25%, >50%, >75%). Post-processing improved agreement on the neoLUSC hold-out set, raising total and 'immature TLS' pixel-level F1 to 0.81 and 0.79 and 'mature TLS' F1 to 0.61.

In lymph nodes, the model reliably captured lymphoid follicles and classified them into primary follicles and germinal center-containing mature follicles, reported as mature secondary lymphoid structures (SLS), as confirmed by expert-pathologist review.

TLS and mature SLS readouts comprised area, count, density, polygon shape descriptors, and, for TLS, inside-TLS as well as surrounding-ring cell composition. For lymph nodes, it included cell composition within annotated lymphoid areas.

**Statistical analysis**

Group differences in each readout, between the therapy-naïve and neoadjuvant cohorts, and for the modality analysis pairwise among the therapy-naïve group and the individual neoadjuvant regimens, were tested with the Mann–Whitney U test, with the rank-biserial correlation r as effect size. Categorical distributions, including the immune-desert/excluded (IDE) phenotype, were compared with Fisher's exact test. Associations between compartments were quantified by Spearman rank correlation (ρ), computed both pooled across cohorts and separately within each cohort. For survival, continuous readouts were standardised and entered individually into univariable Cox proportional-hazards models (hazard ratio per standard-deviation increase), and the clinical covariates (UICC and nodal stage, age, sex, and neoadjuvant status) were tested as univariable predictors in the same way. Kaplan–Meier curves with median-split stratification and the log-rank test provided complementary group-level comparisons. All tests were two-sided, and p-values are reported without correction for multiple comparisons.

## Results

### Neoadjuvant-treated tumors have fewer mature, organised lymphoid structures and lower stromal immune cell densities than therapy-naïve tumors

We first characterised the lymphoid architecture and immune cell composition of treated and therapy-naïve tumors and their draining lymph nodes, comparing TLS and mature SLS, stromal and intratumoral immune-cell densities, as well as the resulting immune phenotype across cohorts (**Fig. 1**).

Representative H&E sections with corresponding AI-model overlays illustrate TLS in tumors and mature SLS in tumor-distal lymph nodes for a naivLUSC and a neoLUSC patient (**Fig. 1A**). TLS and SLS are separated into mature (with germinal center) and immature or primary (without germinal center) ones. In tumor and corresponding draining lymph node samples of a naivLUSC patient the model identifies multiple mature TLS (row 1) and mature SLS (row 3) alongside immature ones, whereas the tumor and lymph node of a neoLUSC patient contain predominantly immature TLS (row 2) and SLS (row 4) with no mature counterparts. The total TLS density did not differ between cohorts (**Fig. 1B**). Neither total TLS density (rank-biserial $r = -0.21$, $p = 0.26$) nor immature TLS density ($r = -0.20$, $p = 0.29$) differed between neoLUSC and naivLUSC resections. The difference was instead confined to the mature, organised compartment: Mature TLS density trended lower in neoLUSC tumors ($r = -0.28$, $p = 0.11$), and mature TLS count was significantly lower ($r = -0.36$, $p = 0.034$). The same directional pattern extended into the draining lymph nodes (**Fig. 1C**). Mature SLS area per lymphoid tissue tended to be lower in neoLUSC patients in the tumor-proximal node ($r = -0.38$, $p = 0.051$) and was significantly lower in the tumor-distal node ($r = -0.55$, $p = 0.0033$). In conclusion, the lower mature TLS/SLS readouts in neoLUSC were directionally consistent along the tumor-to-node axis, and within the draining nodes was more pronounced distally than proximally.

The tumor microenvironment was resolved using the HETA models, which segment WSI into tissue compartments and classify individual cells. Representative naivLUSC and neoLUSC tumors are shown as the H&E alongside the corresponding model overlays in **Fig. 1D**. Quantification of the tumor microenvironment revealed directionally consistent lower stromal immune cell densities in neoLUSC compared to naivLUSC tumors, with the largest difference for stromal plasma cells (**Fig. 1E**). Stromal plasma-cell density was 72% lower ($r = -0.54$, $p = 0.0036$) and stromal lymphocyte density 48% lower ($r = -0.45$, $p = 0.015$). Within the carcinoma compartment, plasma-cell density was 53% lower ($r = -0.39$, $p = 0.038$). Myeloid and granulocytic populations trended in the same direction without reaching significance (macrophage $r = -0.22$, $p = 0.24$; granulocyte $r = -0.27$, $p = 0.15$). Despite these immune cell differences, overall tissue architecture remained broadly comparable between arms. The stroma area was modestly larger (+10%) and vessel area smaller (−33%, $r = -0.34$, $p = 0.068$), while carcinoma and necrosis fractions were essentially stable. The smaller vessel area was one potential structural difference, which did not reach significance. But the slides remained otherwise comparable, suggesting that the lower immune cell densities were not explained by gross differences in tissue composition.

In the next step tumors were assigned an immune phenotype using a density-based inflamed–desert–excluded (IDE) classifier (**Fig. 1F**). Each tumor was positioned by its lymphocyte density in two compartments, within the carcinoma (x-axis) and within the tumor-associated stroma (y-axis) in the annotated whole tumor region, and classified by thresholds derived from the upper tertiles of these per-slide density distributions (dashed lines):

Tumors with high intratumoral lymphocyte density were assigned an inflamed (hot) phenotype, those with low intratumoral but high stromal lymphocyte density an excluded phenotype, and those low in both a desert (cold) phenotype. Consistent with lower immune cell infiltration, slides of treated tumors more frequently displayed a cold, immune-desert phenotype and less frequently a hot phenotype, although this contingency did not reach significance (Fisher exact test, $p = 0.105$, **Fig. 1G**). Spatial neighborhood analysis reinforced the same picture (**Fig. 1H**): The stromal neighborhood contrast matrix at 40 µm was almost uniformly negative across all cell pairs where lymphocyte or plasma cell was the neighbor, indicating fewer immune cell–cell contacts in neoLUSC tumors. The difference was organised by neighbor rather than by centre cell: Plasma cells were the neighbor with the largest difference across nearly all cell types (lymphocyte-, plasma-, macrophage-, and granulocyte-centred neighborhoods all lower at $p < 0.01$), whereas macrophages and granulocytes were the neighbors with the smallest differences.

**Plasma cell densities correlate across lymphoid sites in both cohorts, whereas their correlation with tumor lymphocytes is observed only in therapy-naïve tumors**

We next examined how these compartments are coordinated with one another, and whether that coordination differs between neoLUSC and naivLUSC patients, by examining the cohort-specific correlation structure of the lymphoid, plasma cell and immune cell density readouts (**Fig. 2**).

Tracing the plasma cell axis from the draining lymph nodes into the tumor showed numerically higher correlations between lymphoid sites in neoLUSC than in naivLUSC patients, whereas the correlation with the tumor lymphocyte infiltrate seen in naivLUSC was not observed in neoLUSC (**Fig. 2A**). Plasma cell density in the tumor-distal and tumor-proximal lymph nodes was correlated in neoLUSC tumors ($\rho = 0.74$) and positively but more weakly correlated in naivLUSC tumors ($\rho = 0.38$). The same was true for the link between tumor-proximal lymph node and peri-TLS plasma cell density (neoLUSC $\rho = 0.72$, $p < 0.001$; naivLUSC $\rho = 0.17$, not significant). The local (peri-TLS) and regional (lymph node) plasma cell compartments therefore co-varied tightly in neoLUSC patients, consistent with a synchronised humoral compartment across lymphoid sites, which appeared to be more pronounced in the neoLUSC group compared to the naivLUSC group. This coordination did not extend to the tumour lymphocyte infiltrate: Peri-TLS plasma cell density covaried with stromal and intratumoral lymphocyte density in naivLUSC tumors ($\rho = 0.41$ and $\rho = 0.54$, the latter $p = 0.017$), but not in neoLUSC tumors ($\rho = -0.22$ and $\rho = 0.07$), suggesting that the humoral and cellular arms tracked together in untreated tumours but not in treated tumours.

Extending this analysis to the full set of readouts, cohort-split Spearman correlation matrices showed that the correlation architecture reorganised after neoadjuvant therapy rather than uniformly weakening (**Fig. 2B**). The plasma cell readouts (tumor-distal lymph node, tumor-proximal lymph node, and peri-TLS ring) were tightly correlated among themselves in both cohorts, even more pronounced in the neoLUSC cohort. This indicates that the plasma cell compartment was internally coordinated across its lymphoid sources in both cohorts. By contrast, the coupling of this plasma cell compartment to the surrounding TME differed between cohorts: Its positive correlations with the stromal immune populations in naivLUSC tumors was not detectable in neoLUSC tumors, mirroring the pairwise pattern in **Fig. 2A**. Most strikingly, the correlation between tumor TLS density and the local plasma cell readouts was weakly positive in naivLUSC tumors and negative in neoLUSC tumors. The stromal immune cells were correlated with one another in both cohorts, and the secondary lymphoid structures were only weakly linked to the other compartments throughout. Tumor TLS density showed a positive correlation with the stromal immune compartment in both cohorts and showed little relation to intratumoral carcinoma lymphocytes, so the therapy-associated decoupling was concentrated in the plasma cell readouts rather than affecting the organised lymphoid compartment as a whole. Taken together, plasma cell densities were correlated across the draining lymph nodes and the peri-TLS ring in both cohorts, and the correlation between the lymph nodes and the peri-TLS ring was stronger in the neoLUSC cohort. In contrast, the association of plasma cells with tumor TLS density and with stromal lymphocytes, present in naivLUSC tumors, was not observed in neoLUSC tumors. The difference between cohorts therefore lies in the link between plasma cells and the TME, not in their link to the draining lymph nodes.

**Treatment-associated differences are directionally consistent across modalities and show concordant, non-significant trends toward longer survival**

Next we assessed whether the treatment-associated differences were consistent across neoadjuvant modalities and whether any of the lymphoid or TME readouts related to overall survival (**Fig. 3**). We first summarised the treatment-associated differences across all readout families using group-level radar plots of the z-normalised readouts, comparing cohorts (**Fig. 3A**) and neoadjuvant modalities (**Fig. 3B**). At the cohort level (**Fig. 3A**), the TME profile of neoLUSC tumors was systematically smaller than that of naivLUSC tumors: NeoLUSC values lay inside the naivLUSC profile on every stromal and intratumoral (carcinoma) axis, most markedly for stromal and intratumoral plasma cells and intratumoral lymphocytes. The tumor TLS and lymph node SLS families, by contrast, differed in both directions rather than being uniformly lower: Lower values were concentrated in the mature and structural axes (mature TLS density, mature SLS area), while other axes were similar or higher, notably peri-TLS ring lymphocyte density and nodal macrophage density, which were higher in neoLUSC. The dominant pattern was therefore broadly lower stromal and intratumoral immune cell densities in neoLUSC alongside mixed-direction differences in the organised lymphoid structures, in which structural and mature readouts were lower while selected cellular axes were similar or higher. Stratifying by modality (**Fig. 3B**) showed that this pattern was shared across regimens but not identical between them. The chemotherapy and chemoradiotherapy groups showed broadly overlapping, inward-shifted TME profiles, while the radiotherapy group showed the most inward-shifted TME profile of all groups. The lymphoid families again showed mixed-direction differences rather than uniformly lower values, and several axes diverged in a modality specific manner: Within the lymph node SLS family, macrophage and plasma cell densities were higher mainly in the radiotherapy and chemoradiotherapy groups but not in the chemotherapy group. Additionally, the radiotherapy group showed a higher peri-TLS ring lymphocyte density than the other groups. Because the radiotherapy (n = 2) and chemoradiotherapy (n = 5) subgroups are small, these group-summarised profiles are descriptive and direction-of-difference only, and individual divergent axes should be read with particular caution.

Modality-stratified per-slide distributions were consistent with this overview (**Fig. 3C–F**). The proportion of cold tumors was progressively higher from naivLUSC tumors through the chemotherapy and chemoradiotherapy groups, with both radiotherapy cases classified as cold (**Fig. 3C**). Total TLS density did not differ significantly between any groups (**Fig. 3D**), and mature TLS density was higher in naivLUSC tumors at trend level only (chemotherapy vs NaivLUSC p = 0.065; **Fig. 3E**). Mature SLS area showed the clearest separation, being significantly lower than naivLUSC in both, the chemotherapy (p = 0.018) and chemoradiotherapy (p = 0.007), groups (**Fig. 3F**). Where present, the significant contrasts were between a treated modality and the therapy-naïve group; no pairwise comparison between neoadjuvant modalities reached significance.

Survival analysis in the discovery cohort was limited by sample size (n = 32 evaluable, 21 events), and no univariable readout reached statistical significance. The direction of association was nonetheless consistent: The stromal immune compartments, lower in neoLUSC tumors, each showed a protective tendency, with univariable hazard ratios (HR) below 1 for stromal lymphocyte density (HR 0.84), stromal plasma cell density (HR 0.88), stromal macrophage density (HR 0.81), and lymphocyte–plasma cell co-localization (HR 0.84) (**Fig. 3G**). The clearest tendencies were seen for the organised lymphoid readouts: By Kaplan–Meier median split, higher TLS density (median overall survival 1428 vs 749 days; log-rank p = 0.069; **Fig. 3I**) and higher peri-TLS ring plasma cell density (924 vs 471 days; p = 0.129; **Fig. 3J**) were each associated with longer survival. The cohort split itself was null (HR 0.86, p = 0.74; **Fig. 3H**).

**An independent validation cohort confirms the stromal immune signature and supports its prognostic value**

To validate these findings in an adequately powered, independent setting, we applied the cell level readouts to an independent validation cohort of 458 LUSC patients (46 neoadjuvant-treated, 412 treatment-naïve; **Fig. 4**). The cohort contrast reproduced the stromal immune signature identified in the discovery cohort (**Fig. 4A**): Stromal lymphocyte density, stromal plasma cell density, and stromal lymphocyte–plasma cell co-localization were each significantly lower in neoadjuvant-treated tumors (rank-biserial r = −0.18, −0.22, and −0.18; p = 0.044, 0.014, and 0.041, respectively). In contrast, stromal macrophage density did not differ significantly between groups (p = 0.22), recapitulating the absence of a significant macrophage difference seen in the discovery cohort. As expected for a neoadjuvant-treated subgroup, treated patients presented at higher UICC and nodal stage (both p < 0.001).

In this larger cohort, the survival associations that were directionally suggestive, but underpowered in the discovery set, reached significance (**Fig. 4B**). In univariable Cox regression (n = 458, 255 events), higher densities of the immune compartments that were lower in treated tumors were each associated with a lower hazard of death: stromal lymphocyte density (HR 0.77 per +1 SD, p = 0.005), stromal plasma cell density (HR 0.86, p = 0.038), intratumoral lymphocyte density (HR 0.80, p = 0.027), and both stromal co-localization measures (HR 0.83, p = 0.010 and 0.021) were associated with longer overall survival, while stromal macrophage density was not (HR 0.98, p = 0.80). Consistent with the higher stage of the treated subgroup, neoadjuvant status itself was associated with shorter survival (HR 1.50, p = 0.026), as were UICC and lymph node stage.

The spatial cell-based prognostic associations of the TME were confirmed by Kaplan–Meier analysis, which additionally revealed differences in the magnitude and structure of the separations (**Fig. 4D–F**). Higher stromal plasma cell density and higher stromal lymphocyte–plasma cell co-localization were both associated with substantially longer survival, with median overall survival differences of 26 months (67 vs 41 months; log-rank p = 0.0036; **Fig. 4E**) and 30 months (66 vs 36 months; p = 0.0010; **Fig. 4F)**, respectively. In both, the survival curves diverged progressively over the 60-month horizon rather than separating early. In contrast, stromal lymphocyte density alone separated the groups only weakly and non-significantly (58 vs 50 months; p = 0.156; **Fig. 4D**). The three markers thus formed a clear rank order: The spatial lymphocyte–plasma cell co-localization and the plasma cell density were the strongest discriminators, whereas bulk stromal lymphocyte density was the weakest. Notably, stromal lymphocytes carried little prognostic information on their own but became the strongest stratifier when their spatial co-localization with plasma cells was considered, suggesting that the prognostic signal is more closely related to the spatially coupled humoral compartment than to lymphocyte abundance per se. TLS prognostic associations were not validated in the larger cohort, as it is a TMA-based cohort.

## Discussion

Chemotherapy, chemoradiation and radiotherapy are widely expected to stimulate antitumor immunity, yet their capacity to deplete lymphocytes and disrupt lymphoid architecture is equally well documented. How these opposing effects resolve in the treated tumor and its draining nodes remains largely unquantified. We therefore examined the local and regional immune response in matched therapy-naïve and neoadjuvant-treated lung squamous cell carcinoma, using whole-slide AI-based tissue segmentation (including tertiary lymphoid structure detection) and cell classification to resolve the tumor immune microenvironment and lymphoid architecture along a tumor-to-lymph node axis. This approach captures not only cellular composition but also the spatial organisation of the response. Against the expectation of net immune activation, treated tumors did not differ from therapy-naïve tumors in a simple manner: Rather than uniformly higher or lower local immune readouts, the differences were selective and spatially organised, in which mature, germinal-centre-bearing TLS and the stromal plasma cell compartment were lower while other cellular compartments differed less. This pattern held across the tumor and its draining lymph nodes and on cell composition level was reproduced in an independent validation cohort. Treated tumors thus differed in the maturation state of the lymphoid response rather than lacking it, with the largest differences in the germinal-centre-associated humoral arm.

Myeloid cells showed a different pattern. Macrophage density was not significantly lower in treated patients in any cohort or compartment. The only difference in the opposite direction was a higher macrophage density in the tumor-draining lymph nodes of irradiated patients, while granulocyte density did not differ significantly. We hypothesise that this pattern, and the higher macrophage density in particular, may reflect a repair response to treatment rather than an intrinsic resistance of the myeloid cells to therapy. Cytotoxic treatment destroys tissue and kills cells on a large scale, and the recruitment of macrophages to clear cellular debris and remodel the injured tumor bed is a well-recognised consequence [25].

The most informative differences emerged not from any single compartment but from the relationships between them. In neoLUSC patients, the plasma-cell compartment was more tightly correlated across lymphoid sites: Densities in the distal and proximal draining nodes, and between the proximal node and the peritumoral TLS ring, co-varied more closely than in untreated tumors. In contrast, the coupling between the lymphoid output and the effector infiltrate of the tumor was not present: The co-variation between peritumoral plasma cells and the lymphocytes populating the tumor stroma and carcinoma, present in therapy-naïve tumors, was not detectable in treated tumors. Because H&E resolves a single lymphocyte class and does not capture antibody output, these associations reflect co-variation of the humoral and cellular infiltrates rather than a direct effector relationship. The same pattern was visible across the wider correlation structure, where within-compartment coherence was retained in both cohorts while the cross-compartment coupling between the lymphoid–plasma cell axis and the stromal and carcinoma compartments was weaker in neoLUSC tumors. This points to a disruption of the link between the local and systemic immune response rather than a general loss of immunity. A plausible interpretation is that the systemic, node-based arm of the humoral response remains organised and internally coordinated, while the local step, the integration of this output into the TME, is selectively disrupted. This would be consistent with the parallel reduction in mature, germinal-centre-bearing TLS: With the peritumoral maturation machinery impaired, the humoral compartment may remain quantitatively present and systemically synchronised yet functionally disconnected from the tumor it surrounds. That mature TLS decline under cytotoxic neoadjuvant therapy is, in itself, expected, germinal-centre content defines the prognostically relevant TLS fraction in this disease and has been shown to fall in chemotherapy- and radiotherapy-treated squamous lung tumors [8,19,20]. Our data add information on coordination rather than abundance alone. Previous work has described therapy-associated changes in the abundance, maturation and localization of lymphoid structures. We additionally find that the coordination between these compartments differs between treated and therapy-naïve patients: In treated patients, the systemic humoral axis was more tightly correlated internally, while its correlation with the local effector compartment was not detectable. Along the tumor-to-lymph node axis, the lesion is therefore best understood not as a loss of lymphoid structures or cells but as a decoupling within an otherwise preserved local–systemic circuit.

The lower number of mature lymphoid structures was not uniform across treatment modalities. Across the three compartments, the profiles of treated tumors differed from naïve ones in shape, not solely in size, in the radar plots, treated profiles lay inside the naïve profiles selectively along the mature-lymphoid and plasma-cell axes, the myeloid axes were comparatively similar, and in the irradiated draining lymph nodes the macrophage axis lay outside. This is consistent with selective rather than uniform differences between cohorts. This difference was most pronounced in the radiation-containing regimens and clearest for the germinal-centre-bearing follicles of the draining lymph nodes, consistent with the radiosensitivity of the germinal-centre reaction and radiotherapy's lymphodepleting action [8,19,20]. Statistical support was constrained by subgroup size: Only the contrasts against treatment-naïve tumors reached nominal significance, for chemoradiation and, less strongly, chemotherapy, none surviving correction for multiple testing, and no modality ranked above another. The radiotherapy-alone group showed the largest decreases, but with only two such patients these fell short of significance. The modality data are therefore best read as descriptive: The distinct compartment shapes are compatible with larger differences in the radiation-containing regimens, but the subgroups are too small to attribute them to any single treatment. Whether these differences carry prognostic weight could not be resolved in the discovery cohort, which was underpowered for survival analysis: With only 21 deaths among the evaluable patients, no single immune or structural readout reached significance. The direction was nonetheless consistent, every feature that was lower in treated tumors pointed the protective way, higher densities associating with longer survival, with TLS density and the peritumoral plasma-cell ring the strongest tendencies. We therefore consider this result a consequence of limited power rather than evidence of no effect, and tested it in an independent cohort of 458 patients. There the same features were significantly and protectively prognostic: Stromal lymphocyte density, stromal plasma cell density, and the co-localization of lymphocytes with plasma cells each predicted longer overall survival. Lymphocyte density was thus prognostic in its own right, in keeping with the established value of tumor-infiltrating lymphocytes [6]. Its co-localization with plasma cells, however, was the more robust discriminator, separating survival groups where lymphocyte density alone did so only weakly. This may reflect a functional, spatially organised facet of the humoral response, the productive proximity of the two compartments, that abundance alone does not fully capture, and one that is difficult to resolve without automated whole-slide quantification. Furthermore the long-term survival benefit associated with high stromal plasma cell density may reflect sustained output from long-lived, affinity-matured plasma cells generated in germinal-centres [26]. In NSCLC, germinal-centre activity within TLS is linked to local plasma cell generation and favourable long-term survival [9], consistent with this interpretation. We did not assess clonal history or antibody affinity, so the germinal-centre origin remains hypothetical. Macrophage density, which did not differ significantly between treated and therapy-naïve patients, was not prognostic, reinforcing the cross-cohort consistency of the myeloid pattern. Neoadjuvant status itself was associated with shorter survival, which we attribute to the higher stage of the treated group, rather than to the immune differences themselves. These findings gain particular significance from how the treatment of resectable disease has since changed. The patients studied here predate immunotherapy in the neoadjuvant setting. Today, neoadjuvant or perioperative platinum chemotherapy combined with a PD-1/PD-L1 inhibitor is standard of care for resectable NSCLC [18]. Because it is the mature, germinal-centre-bearing TLS and B-cell response that have been linked to checkpoint-inhibitor benefit independently of PD-L1 expression [9–11], our finding that this architecture is selectively less prominent in tumors treated with cytotoxic therapy raises a clinically relevant hypothesis: Prior chemoradiation may deplete the immune substrate on which subsequent immunotherapy depends, with implications for how the two modalities are sequenced and timed. This hypothesis should be interpreted with caution because the literature describes a dual effect of therapy. Radiotherapy can also prime antitumor immunity through immunogenic cell death and type-I interferon signalling, and prior radiation has been linked to improved checkpoint-inhibitor outcomes [27]. The net effect is therefore unlikely to be uniform, depending on modality, dose, and the sequencing and timing of the therapies combined. Our data directly support a more limited conclusion: The stromal plasma-cell compartment, and above all its spatial coupling with tumor lymphocytes, is both lower in tumors treated with cytotoxic therapy and prognostic for survival in the external cohort, a combination that makes it a quantitative, whole-slide-measurable index of the humoral response's integrity in the residual tumor. Whether this index also reflects the immune substrate available to subsequent immunotherapy remains an open question which would require prospective, immunotherapy-treated cohorts.

Our conclusions are also limited by the timing of sampling. The specimens capture a single, early time point, tissue resected only weeks after therapy, in the window of maximal cytotoxic injury and early repair rather than long-term immune reconstitution. This may explain why this snapshot emphasises disorganisation where longer-term studies have described radiation-induced lymphoid neogenesis. Within this limitation, our data suggest that neoadjuvant cytotoxic therapy does not globally suppress antitumor immunity. Instead, it selectively disrupts an otherwise preserved loco-regional humoral response, at the final step where this response reaches the tumor.

**Table 1: Clinical and pathological parameters of the discovery cohort separated by neoadjuvant treatment**

| LUSC Cohorts | neoLUSC (n=20) | naiveLUSC (n=20) |
|---|---|---|
| Mean age (y) [SD] | 61.2 [7.1] | 69.1 [7.1] |
| Gender(%) | | |
| m | 15 (75.0) | 15 (75.0) |
| T-Stage (%) | | |
| 1c | 1 (5.0) | 2 (10.0) |
| 2a | 3 (15.0) | 3 (15.0) |
| 2b | 6 (30.0) | 6 (30.0) |
| 3 | 6 (30.0 | 6 (30.0) |
| 4 | 4 (20.0) | 3 (15.0) |
| N-Stage (%) | | |
| 0 | 5 (25.0) | 10 (50.0) |
| 1 | 10 (50.0) | 8 (40.0) |
| 2 | 5 (25.0 | 2 (10.0) |
| M-Stage (%) | | |
| 0 | 20 (100) | 19 (95.5) |
| 1a | 0 | 1 (5.0) |
| R-Status (%) | | |
| 0 | 17 (85.0) | 19 (95.5) |
| 1 | 3 (15.0) | 1 (5.0) |
| UICC (%) | | |
| I | 0 | 3 (15) |
| II | 9 (45.0) | 9 (45.0) |
| III | 11 (55.0) | 7 (35.0) |
| IV | 0 | 1 (5.0) |
| Radiation (%) | 2 (10) | - |
| Chemo(%) | 11 (55.0) | - |
| CRT (%) | 5 (25.0) | - |
| Unclear (%) | 2 (10) | - |

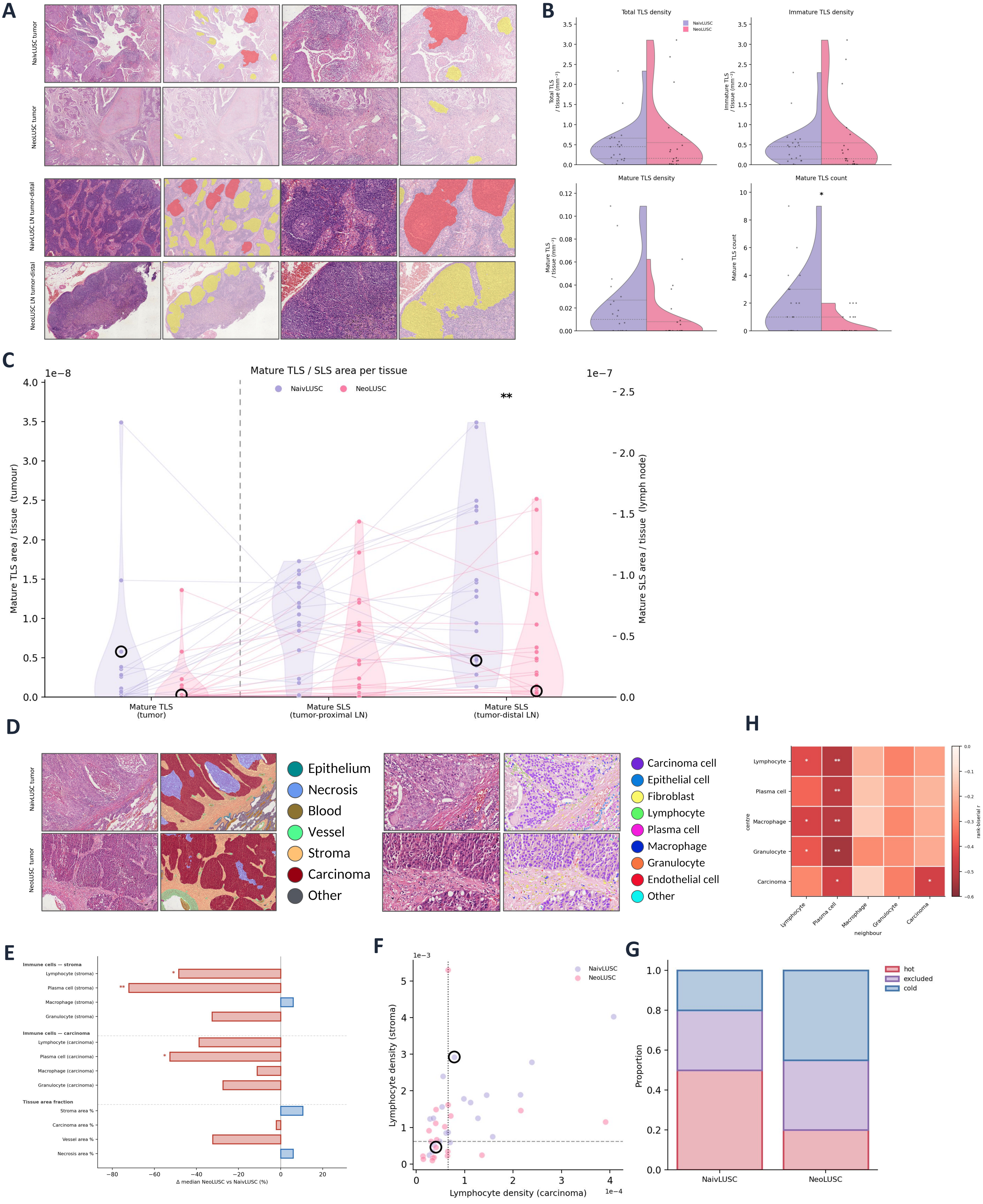
A
NaivLUSC tumor
NeoLUSC tumor
NaivLUSC LN tumor-distal
NeoLUSC LN tumor-distal
B
Total TLS density
NaivLUSC
NeoLUSC
Total TLS / tissue (mm⁻²)
Immature TLS density
Immature TLS / tissue (mm⁻²)
Mature TLS density
Mature TLS / tissue (mm⁻²)
Mature TLS count
*
C
Mature TLS / SLS area per tissue
1e−8
1e−7
NaivLUSC
NeoLUSC
**
Mature TLS area / tissue (tumour)
Mature SLS area / tissue (lymph node)
Mature TLS (tumor)
Mature SLS (tumor-proximal LN)
Mature SLS (tumor-distal LN)
D
NaivLUSC tumor
NeoLUSC tumor
Epithelium
Necrosis
Blood
Vessel
Stroma
Carcinoma
Other
Carcinoma cell
Epithelial cell
Fibroblast
Lymphocyte
Plasma cell
Macrophage
Granulocyte
Endothelial cell
Other
H
Lymphocyte
Plasma cell
Macrophage
Granulocyte
Carcinoma
centre
neighbour
rank-biserial r
E
Immune cells — stroma
Lymphocyte (stroma)
Plasma cell (stroma)
Macrophage (stroma)
Granulocyte (stroma)
Immune cells — carcinoma
Lymphocyte (carcinoma)
Plasma cell (carcinoma)
Macrophage (carcinoma)
Granulocyte (carcinoma)
Tissue area fraction
Stroma area %
Carcinoma area %
Vessel area %
Necrosis area %
Δ median NeoLUSC vs NaivLUSC (%)
F
1e−3
NaivLUSC
NeoLUSC
Lymphocyte density (stroma)
Lymphocyte density (carcinoma)
1e−4
G
hot
excluded
cold
Proportion
NaivLUSC
NeoLUSC

**Figure 1: Lymphoid structures, immune-cell composition, and microenvironment phenotype in treatment-naive (naivLUSC) and neoadjuvant-treated (neoLUSC) lung squamous cell carcinoma.**

**(A)** Representative H&E regions with tertiary lymphoid structures (TLS) in tumors (rows 1-2, naivLUSC and neoLUSC) and lymphoid follicles (secondary lymphoid structures, SLS, rows 3-4, naivLUSC and neoLUSC) in lymph nodes. Model overlays highlight immature TLS and primary SLS in yellow and mature TLS and SLS with germinal centers in red.

**(B)** Per-slide tumor TLS quantification by cohort: Total TLS density, immature TLS density, mature TLS density (per $mm^2$ tissue), and mature TLS count.

**(C)** Mature lymphoid area per tissue along the tumor-to-node axis. Mature TLS in tumor (left axis) and mature SLS in tumor-proximal and tumor-distal lymph nodes (right axis) shown as split violins with per-patient connectors. Black circles mark samples shown in **A**.

**(D)** Representative tumor H&E regions (row 1 naivLUSC, row 2 neoLUSC) with H&E, tissue segmentation, and cell detection/classification at higher magnification.

**(E)** Change in tumor micorenvironment (TME) composition (Δ median neoLUSC versus naivLUSC, %) for stromal and intratumoral immune-cell densities (lymphocyte, plasma cell, macrophage, granulocyte) and tissue-area fractions (stroma, carcinoma, vessel, necrosis). Bar direction indicates relative decrease (red) or increase (blue) in neoLUSC tumors compared to naivLUSC tumors.

**(F)** Density-based immune-phenotype (inflamed, desert, excluded, IDE) assignment from per-slide stromal versus intratumoral (carcinoma) lymphocyte density. Dashed lines indicate classification thresholds.

**(G)** Proportion of immune phenotypes (hot, excluded, cold) per cohort.

**(H)** Differential stromal cell–cell co-localization within a 40 μm neighborhood (rank-biserial effect size, neoLUSC versus naivLUSC) for each center–neighbour cell pair.

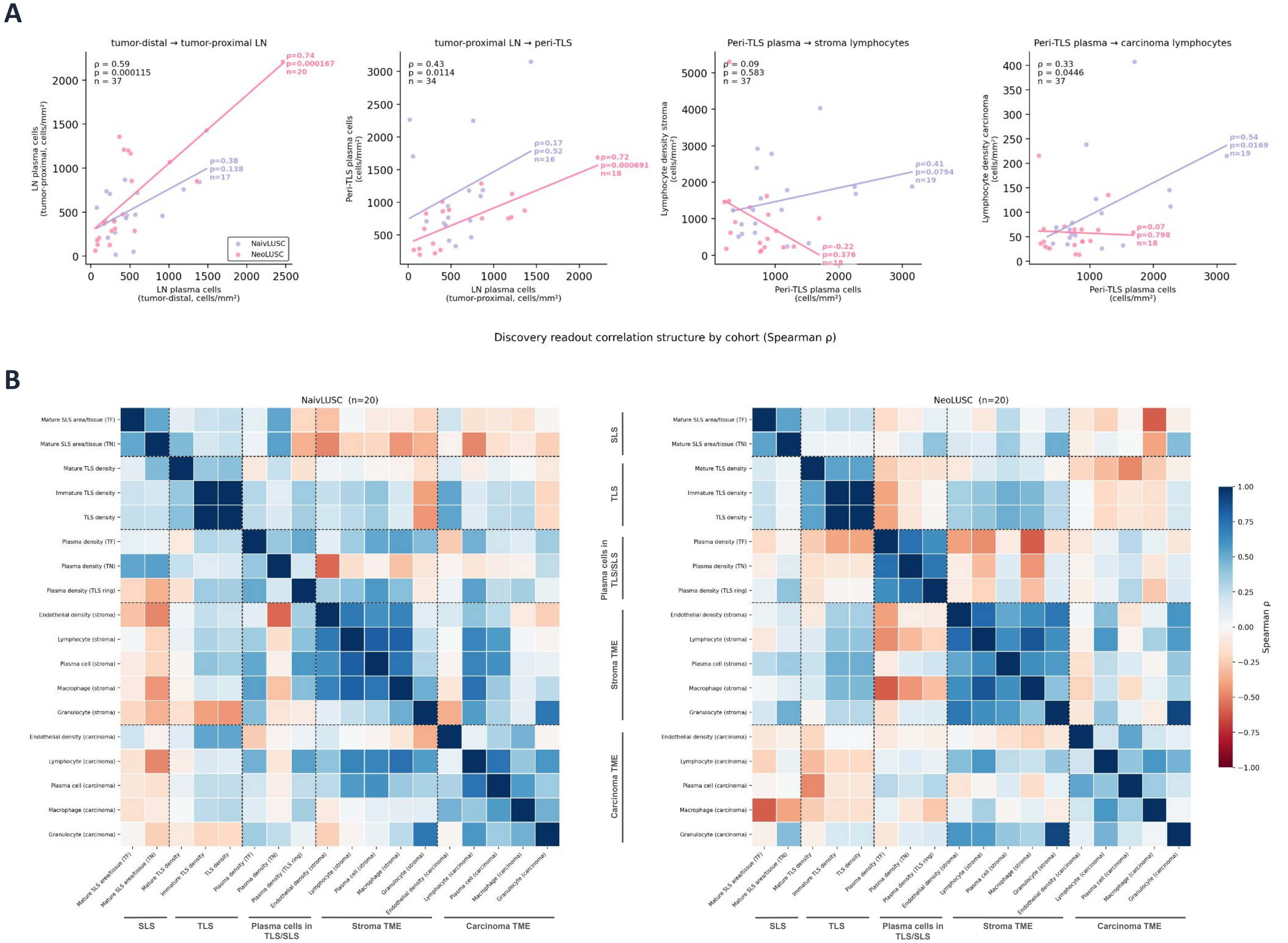

**Figure 2: Cohort-specific correlation structure of plasma cell, tertiary lymphoid structure (TLS), and immune cell density readouts in naivLUSC and neoLUSC patients.**

**(A)** Pairwise Spearman correlations between per tissue normalized readouts, shown as scatter plots with per cohort linear fits (naivLUSC, purple; neoLUSC, pink); each panel is annotated with the pooled and cohort specific correlation coefficient (ρ), p value, and sample size (n) for the following compartment correlations: Plasma cell density in tumor-distal versus tumor-proximal lymph nodes; tumor-proximal lymph node versus peri-TLS plasma cell density, peri-TLS plasma cell density versus stromal lymphocyte density; and peri-TLS plasma cell density versus intratumoral (carcinoma) lymphocyte density. Correlations involving peri-TLS plasma are restricted to TLS-bearing tumours (n = 19 naivLUSC, 18 neoLUSC). The lymph node–ring correlation is further restricted to cases with an evaluable tumor-proximal lymph node (n = 16 naivLUSC, 18 neoLUSC).

**(B)** Cohort-specific Spearman correlation matrices of the per-tissue–normalized readouts, computed separately within each cohort (naivLUSC, n = 20; neoLUSC, n = 20).

The 18 readouts are arranged into five blocks (dashed separators), in axis order: Mature SLS (SLS area per annotated lymphoid tissue in tumor-near (TN) and tumor-far (TF) lymph node), TLS (mature, immature, and total TLS density), plasma cells in TLS/SLS (plasma cell density tumor-far lymph node, tumor-near lymph node, and peri-TLS ring), stromal TME (endothelial cell, lymphocyte, plasma cell, macrophage, and granulocyte density in stroma) and carcinoma TME (the same five cell densities within carcinoma).

Cell colour encodes the Spearman correlation coefficient ρ from −1 (red) to +1 (blue), as indicated by the colour bar. Correlations are exploratory and not corrected for multiple comparisons.

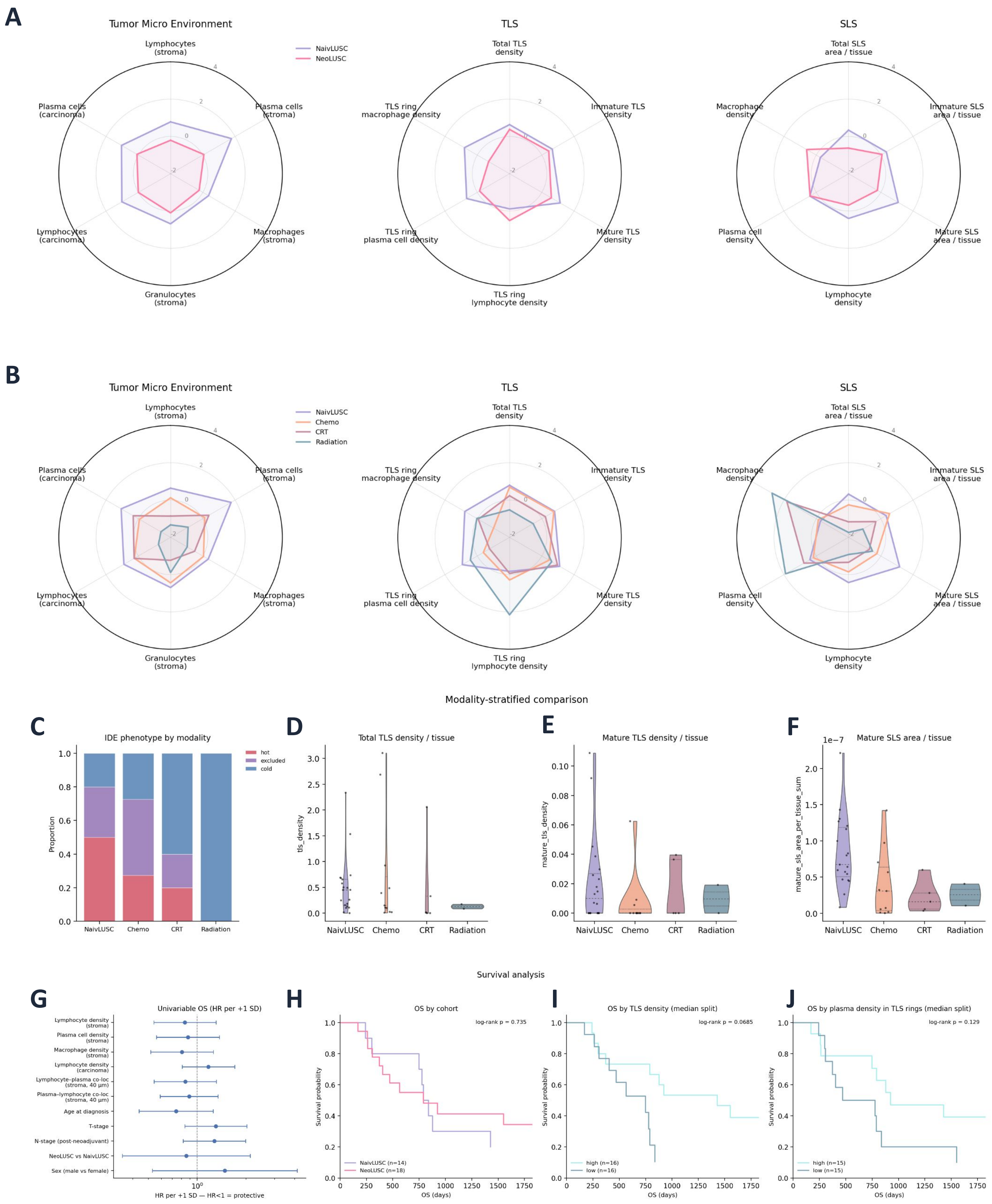
A
Tumor Micro Environment
TLS
SLS
NaivLUSC
NeoLUSC
Lymphocytes (stroma)
Plasma cells (stroma)
Macrophages (stroma)
Granulocytes (stroma)
Lymphocytes (carcinoma)
Plasma cells (carcinoma)
Total TLS density
Immature TLS density
Mature TLS density
TLS ring lymphocyte density
TLS ring plasma cell density
TLS ring macrophage density
Total SLS area / tissue
Immature SLS area / tissue
Mature SLS area / tissue
Lymphocyte density
Plasma cell density
Macrophage density
B
Tumor Micro Environment
TLS
SLS
NaivLUSC
Chemo
CRT
Radiation
Modality-stratified comparison
C
IDE phenotype by modality
hot
excluded
cold
Proportion
NaivLUSC
Chemo
CRT
Radiation
D
Total TLS density / tissue
tls_density
E
Mature TLS density / tissue
mature_tls_density
F
Mature SLS area / tissue
1e−7
mature_sls_area_per_tissue_sum
Survival analysis
G
Univariable OS (HR per +1 SD)
Lymphocyte density (stroma)
Plasma cell density (stroma)
Macrophage density (stroma)
Lymphocyte density (carcinoma)
Lymphocyte–plasma co-loc (stroma, 40 μm)
Plasma–lymphocyte co-loc (stroma, 40 μm)
Age at diagnosis
T-stage
N-stage (post-neoadjuvant)
NeoLUSC vs NaivLUSC
Sex (male vs female)
HR per +1 SD — HR<1 = protective
H
OS by cohort
log-rank p = 0.735
NaivLUSC (n=14)
NeoLUSC (n=18)
Survival probability
OS (days)
I
OS by TLS density (median split)
log-rank p = 0.0685
high (n=16)
low (n=16)
J
OS by plasma density in TLS rings (median split)
log-rank p = 0.129
high (n=15)
low (n=15)

**Figure 3: Treatment-associated changes in lymphoid and TME readouts across cohorts and neoadjuvant modalities, and their relationship to overall survival**

**(A)** Cohort comparison of group-summarized, normalized readouts for three readout families, TME (stromal and intratumoral immune cell densities), tumor TLS (total, immature, and mature TLS density and peri-TLS ring macrophage, plasma cell, and lymphocyte density), and secondary lymphoid structures (immature and mature SLS area per lymphoid tissue and lymphoid tissue immune cell densities). NaivLUSC, purple, neoLUSC, pink. **(B)** The same readout families and axes stratified by neoadjuvant modality (naivLUSC, purple; chemotherapy, orange; chemoradiotherapy [CRT], pink; radiotherapy, green). **(C)** Immune phenotype (IDE) proportions (hot, excluded, cold) by neoadjuvant modality (naivLUSC, chemotherapy, CRT, radiotherapy).

**(D–F)** Per-slide distributions by modality (violin plots with individual values and median/quartile lines): **(D)** Total TLS density per tissue, **(E)** mature TLS density per tissue, and **(F)** mature SLS area per lymphoid tissue. **(G)** Univariable overall survival analysis (Cox proportional-hazards; hazard ratio per +1 SD with 95% confidence interval) for stromal lymphocyte, plasma cell, and macrophage density, intratumoral (carcinoma) lymphocyte density, lymphocyte–plasma cell and plasma cell–lymphocyte colocalization (stroma, 40 µm), age at diagnosis, T stage, post-neoadjuvant N stage, cohort (neoLUSC vs naivLUSC), and sex (male vs female). HR $< 1$ indicates a protective association.

**(H–J)** Kaplan–Meier overall survival curves with log-rank *P* values: **(H)** by cohort (naivLUSC, n = 14, neoLUSC, n = 18), **(I)** by tumor TLS density (median split: high, n = 16; low, n = 16), and **(J)** by peri-TLS ring plasma cell density (median split: high, n = 15; low, n = 15).

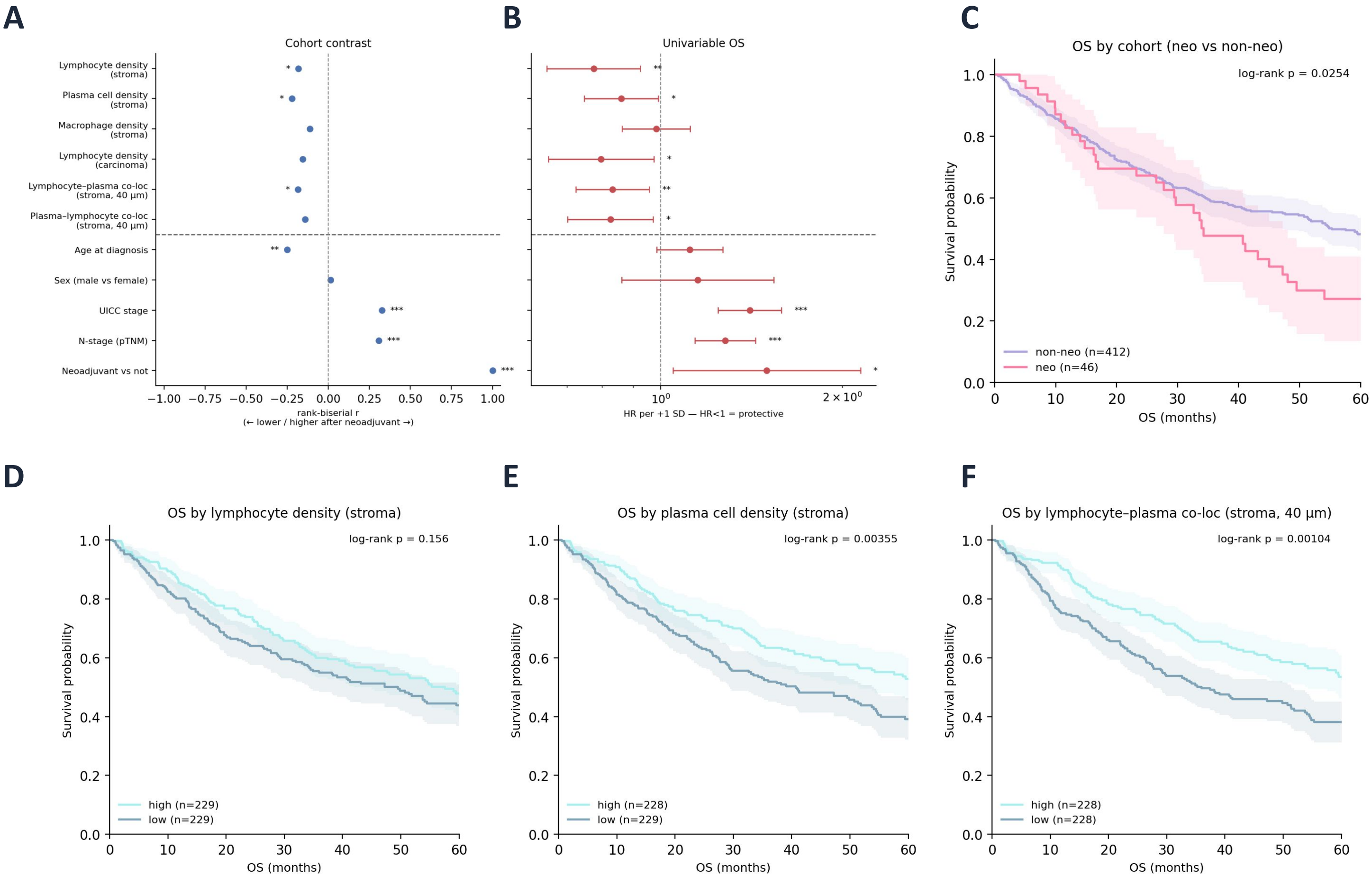

**Figure 4: Independent confirmation of the neoadjuvant-associated stromal immune signature and its prognostic relevance in a large external LUSC cohort.**
**(A)** Cohort contrast (neoadjuvant (n=46) vs non-neoadjuvant (n=412)) for stromal lymphocyte, plasma cell, and macrophage density; intratumoral (carcinoma) lymphocyte density; lymphocyte–plasma cell and plasma cell–lymphocyte colocalization (stroma, 40 µm) in stroma; and clinicopathologic variables (age, sex, UICC stage, pathologic N stage [pTNM], and neoadjuvant status), expressed as rank biserial effect size. Negative values indicate lower values after neoadjuvant therapy. **(B)** Univariable overall survival analysis for the same variables (Cox proportional hazards; hazard ratio per +1 SD with 95% confidence interval). HR < 1 indicates a protective association. **(C)** Kaplan–Meier overall survival by cohort (non-neoadjuvant, n = 412; neoadjuvant, n = 46).
**(D–F)** Kaplan–Meier overall survival: **(D)** by median split of stromal lymphocyte density, **(E)** by stromal plasma-cell density, and **(F)** by stromal lymphocyte–plasma cell colocalization (40 µm). All Kaplan–Meier comparisons show log-rank p values. Overall survival is shown in months over a 60-month horizon.